\documentclass[fleqn,12pt]{wlscirep}
\usepackage[utf8]{inputenc}
\usepackage[T1]{fontenc}
\usepackage{multirow}

\usepackage{graphicx}
\usepackage{url}
\usepackage{amsthm}
\usepackage{amsfonts}
\usepackage{amsmath}
\usepackage{mathrsfs}
\usepackage{algorithm}
\usepackage{algorithmic}

\newcommand{\alg}{\textsc{Data Shapley }}
\newtheorem*{prop}{Proposition}

\begin{document}

\title{What is your data worth? Equitable Valuation of Data}

\author[1]{Amirata Ghorbani}
\author[2,3,*]{James Y. Zou}

\affil[1]{Department Electrical Engineering, Stanford University, CA, USA}
\affil[2]{Department of Bioemdical Data Science, Stanford University, CA, USA}
\affil[3]{Chan-–Zuckerberg Biohub, San Francisco, CA, USA}

\affil[*]{corresponding author: jamesz@stanford.edu}
\maketitle

\section*{Abstract}
As data becomes the fuel driving technological and economic growth, a fundamental challenge is how to quantify the value of data in algorithmic predictions and decisions. 
For example, in healthcare and consumer markets, it has been suggested that individuals should be compensated for the data that they generate, but it is not clear what is an equitable valuation for individual data.
In this work, we develop a principled framework to address data valuation in the context of supervised machine learning. Given a learning algorithm trained on $n$ data points to produce a predictor, we propose data Shapley as a metric to quantify the value of each training datum to the predictor performance. Data shapley value uniquely satisfies several natural properties of equitable data valuation. We develop Monte Carlo and gradient-based methods to efficiently estimate data Shapley values in practical settings where complex learning algorithms, including neural networks, are trained on large datasets. In addition to being equitable, extensive experiments across biomedical, image and synthetic data demonstrate that data Shapley has several other benefits: 1) it is more powerful than the popular leave-one-out or leverage score in providing insight on what data is more valuable for a given learning task; 2)  low Shapley value data effectively capture outliers and corruptions; 3)  high Shapley value data inform what type of new data to acquire to improve the predictor.

\paragraph{Introduction}    
    Data is the fuel powering artificial intelligence and therefore it has value. Various sectors such as health-care and advertising are becoming more and more dependent on the data generated by individuals; a dependence similar to how they depend on labor and capital\cite{posner2018radical}. As the legal system moves toward recognizing individual data as property~\cite{regulation2018general}, a natural problem to solve is to equitably assign value to this property. The result is the ability to have fair compensation individuals' data. ''\alg'' is designed to provide such valuation in one of the widely-used settings of data usage, namely machine learning.
    
    Throughout this work, we focus on valuation of data in the setting of supervised learning as one of the main pillars of artificial intelligence. In supervised machine learning, the data is in the form of input-output pairs; e.g. location and temperature, customers search history and their shopping list, an object's image and its name, and so forth. Using a learning algorithm, a predictive model is then trained to predict the output given the input. The learning algorithm takes a set of train data points ($n$ input-output pairs $\{(x_i, y_i)\}_{i=1}^n$) as its input and outputs the learned predictive model. To assess the quality of the trained model, a performance metric identifies how useful the trained model is. For instance, given a set of input-output pairs of data different from that of the training set, what percentage of outputs are correctly predicted given the inputs. As mentioned, supervised machine learning consists of three ingredients" training data, learning algorithm, and performance metric. As a consequence, to make sense of value for data, we need the same ingredients in our investigation: we are given the fixed data set used for training the machine learning model, the algorithm that trained the model, and the metric to assess the learned model's performance.
    
    Given the three ingredients, we want to investigate two questions: 1) what is an equitable measure of the value of each train datum $(x_i, y_i)$ to the learning algorithm $\mathcal{A}$ with respect to the performance metric $V$; and 2) how do we efficiently compute this data value in practical supervised learning settings. For example, assume the train data comes from $N=1000$ patients where each patient has provided their medical history and whether they were diagnosed with a heart disease. The learning algorithm uses this data to train a small neural network model and we assess the performance of the trained classifier by its accuracy on a test set; a separate set of patients whose data is not used for training the model. How can we compensate each patient?
 
     It is important to note that assigning value to data in the setting of supervised learning is not equal to assigning a universal to it. Instead, the value of each datum  depend on the three pillars of supervised learning: learning algorithm, the performance metric as well as on the rest of the train data. This dependency is reasonable and desirable in machine learning. Certain data points could be more important if we are training a logistic regression model instead of a neural network model. Similarly, if the performance metric changes---e.g. regressing to the age of heart disease onset instead of heart disease incidence---then the value of a certain patient's data should change if the patient data was more important for one task. 
 

    \begin{figure}
        \centering
        \includegraphics[width=0.65\linewidth]{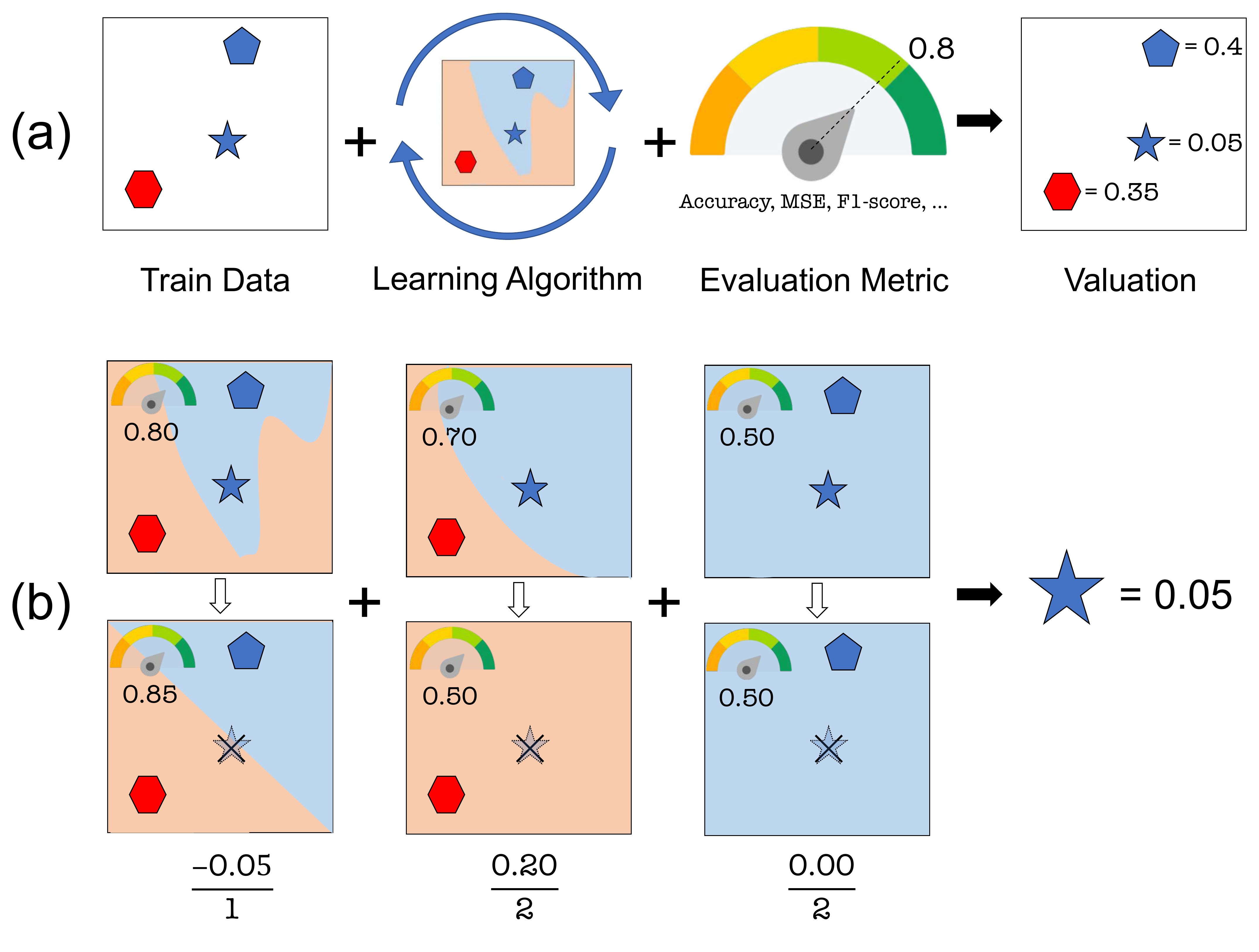} 
        \caption{(a) Schematic overview of data valuation for supervised learning. A set of individual data points are used as the input of a learning algorithm and outputs a trained model the performance of which is measured on a pre-defined set of test data points. 
        (b) For each piece of data, its value is a weighted average of how much it contributes to a subset of rest of the data sources. The weights are determined by Eqn. ~\ref{eq:shapley}.
        \label{fig:Schematic}}
    \end{figure}

\paragraph{Overview} 
    \alg introduces a natural formulation for the problem of equitable data valuation in supervised machine learning. The supervised machine learning has three main ingredients. The first ingredient is the training set. Let $D = \{(x_i, y_i)\}_1^{n}$ be our fixed training set. $D$ comes from $n$ different sources of data where $(x_i, y_i)$ is the $i$'th source. Each data source could be a single input-output pair or a set of pairs. We do not make any distributional assumptions about $D$ including being independent and identically distributed. The outputs ($y_i$'s) can be categorical for the case of classification task or real for the case of regression. The second ingredient is the learning algorithm which we denote it by $\mathcal{A}$. We view $\mathcal{A}$ as a black-box that takes the training data set $D$ as input and returns a predictor. 
    The last ingredient evaluates the learned predictor. The performance score $V$ is another black-box oracle that takes as input any predictor and returns its score. We write $V(S, \mathcal{A})$, or just $V(S)$ for short, to denote the performance score of the predictor trained on train data $S$ using the learning algorithm $\mathcal{A}$. For each data source $(x_i, y_i) \in D$, our goal is to compute its \textbf{data value}, denoted by $\phi_i(D, \mathcal{A}, V) \in \mathbb{R}$, as a function of the three ingredients $D, \mathcal{A}$ and $V$. We will often write it as $\phi_i(V)$ or just $\phi_i$ to simplify notation. We simplify the notation even more for $S$ and $D$ to their set of indices---i.e. $i \in S \subseteq D$ is the same as $(x_i, y_i)$ is in $S \subseteq D$ and therefore $D = \{1, ..., n\}$. 
    
    One simple way to interpret the value of each data source is to measure its contribution to the rest of the train data: $\phi_i = V(D) - V(D - \{i\})$; in other words, how much performance we loose if we remove the $i$'th data source. This method is referred to as the ``leave-one-out'' (LOO) method. The issue, however, is that this valuation scheme does not satisfy the equitable valuation conditions. We believe that an equitable valuation method should satisfy the following conditions:
    
    \begin{enumerate}
    
        \item If $(x_i, y_i)$ does not change the performance if it's added to any subset of the train data sources, then it should be given zero value. More precisely, suppose for all $S \subseteq D-\{i\}$, $V(S) = V(S \cup\{i\})$, then $\phi_i = 0$.
        
        \item If for data $i$ and $j$ and any subset $S \subseteq D - \{i,j\}$, we have $V(S \cup \{i\}) = V(S \cup \{j\})$, then $\phi_i = \phi_j$. In other words, if $i$ and $j$, have exactly equal contribution when added to any subset of our training data, then $i$ and $j$ should be given the same value by symmetry. 
        
        \item When the overall performance score is the sum of separate performance scores, the overall value of a datum should be the sum of its value for each score: $\phi_i(V+W) = \phi_i(V) + \phi_i(W)$ for  performance scores $V$ and $W$. In most ML settings, the model performance is evaluated summing up the individual loss of data points in a test set (0-1 loss, cross-entropy loss, etc): $V =  -\sum_{k \in \mbox{test set}}$  $l_k$ where $l_k$ is the loss of the predictor on the $k$-th test point (the minus means lower loss equal to higher perfroamnce score). Given $V_k = - l_k$ as be the predictor's performance on the $k$-th test point, we can define $\phi_i(V_k)$ to quantify the value of the $i$-th source for predicting the $k$-th test point. If datum $i$ contributes values $\phi_i(V_1)$ and $\phi_i(V_2)$ to the predictions of test points 1 and 2, respectively, then we expect the value of $i$ in predicting both test points---i.e. when $V = V_1 + V_2$---to be $\phi_i(V_1) + \phi_i(V_2)$. 
        
    \end{enumerate}

\begin{prop}
Any data valuation $\phi(D, \mathcal{A}, V)$ that satisfies properties 1-3 above must have the form 
\begin{align}
\label{eq:shapley}
\centering
\phi_i = C \sum_{S \subseteq D - \{i\}}  \frac{V(S \cup \{i\}) - V(S)}{{n - 1 \choose |S|}}
\end{align}
where the sum is over all subsets of $D$ not containing $i$ and $C$ is an arbitrary constant. We call $\phi_i$ the Data Shapley value of source $i$. 
\end{prop}\label{prop:shapley}

\begin{proof}
The expression of $\phi_i$ in Eqn.~\ref{eq:shapley} is the same as the Shapley value defined in game theory, up to the constant $C$ \cite{shapley1953value,shapley1988shapley}. This motivates calling $\phi_i$ the data Shapley value. The proof  also follows directly from the uniqueness of the game theoretic Shapley value, by reducing our problem to a cooperative game \cite{dubey1975uniqueness}. 
 In a cooperative game, there are $n$ players $D=\{1,\dots,n\}$ and a score function $v: 2^{[n]} \rightarrow \mathbb{R}$ assigns a reward to each of $2^n$ subsets of players: $v(S)$ is the reward if the players in subset $S \subseteq D$ cooperate.    
Shapley~\cite{shapley1953value} proposed a scheme to uniquely divides the reward for cooperation of all players $V(D)$ such that each player would get an equaitable share of the reward.  Equity, in his work, is codified by properties that are mathematically equivalent to the three properties that we listed. We view the supervised machine learning problem as a cooperative game: each source in the train data is a player, and the players work together through the learning algorithm $\mathcal{A}$ to achieve prediction score $v(D) = V(D, \mathcal{A})$. \alg computes the equitable share that each player receives from the cooperation. 
\end{proof}

As Eqn.~\ref{eq:shapley} suggests, computing data Shapley value requires exponentially large number of computations with respect to the number of train data sources.

\alg tackles this problem by extending Monte-Carlo approximation methods developed Shapley value ~\cite{mann196large, castro2009polynomial, maleki2013bounding} to data valuation setting: First, we sample a random permutations of sources in the train data. Then, we scan the permutation from the first element to the last element. For each new data source, we add it to the previous sources and compute it improves the performance from only having the previous sources. This marginal contribution of the data source is one monte-carlo sample of its Data Shapley value. Similar to any monte-carloc approximation scheme, we can improve the approximation repeating the same process and taking the average of all marginal contributions of that data source. Additionally, the supervied learning problem offers us another level of approximation. As we scan the sampled permutation and add more and more data sources, the size of the train data is increasing. As the train data size increases, the marginal contribution if adding the next data source becomes smaller and smaller ~\cite{mahajan2018exploring, beleites2013sample}. Therefore, instead of scanning over all of the data sources in the sampled permutation, we truncate the computations once the marginal contributions become small and approximate the marginal contribution of the following elements with zero. We refer to this scheme of approximating Data Shapley as  ``Truncated Monte Carlo Shapley'' (TMC-Shapley) described in Alg.~\ref{alg:TMC-Shapley}. More details of the algorithm are provided in Appendix~\ref{app:TMC-Shapley}. We also introduced a second approximation algorithm tailored for a more specific family of learning algorithms in Appendix~\ref{app:G-shapley}.

        \begin{algorithm}[tb]
          \caption{\textbf{Truncated Monte Carlo Shapley}}
          \label{alg:TMC-Shapley}
              \begin{algorithmic}
                \STATE {\bfseries Input:}  Train data $D = \{1,\dots,n\}$, learning algorithm $\mathcal{A}$, performance score $V$
                \STATE {\bfseries Output:} Shapley value of training points: $\phi_1,\dots,\phi_n$
                \vspace{1mm}
                \STATE Initialize $\phi_i=0$ for $i=1,\dots,n$ and $t=0$
                \WHILE{Convergence criteria not met}
                    \STATE $t \leftarrow t+1$
                    \STATE $\pi^{t}$: Random permutation of train data points
                    \STATE $v^t_0 \leftarrow V(\emptyset, \mathcal{A})$ 
                    \FOR{$j \in \{1,\ldots,n\}$}
                        \IF{$|V(D) - v^t_{j-1}| < \mbox{Performance Tolerance}$}
                            \STATE $v^t_{j} = v^t_{j-1}$
                        \ELSE
                            \vspace{0.5mm}
                            \STATE $v^t_j \leftarrow V(\{\pi^{t}[1],\dots,\pi^{t}[j]\}, \mathcal{A})$
                            \vspace{0.5mm}
                        \ENDIF
                        \STATE $\phi_{\pi^t[j]} \leftarrow \frac{t-1}{t}\phi_{\pi^{t-1}[j]} + \frac{1}{t}(v^t_{j} - v^t_{j-1}) $  
                        \vspace{1mm}
                     \ENDFOR
                \ENDWHILE
                \vspace{1mm}
              \end{algorithmic}
        \end{algorithm}
\section*{Applications of \alg}

In this section, we introduce two main potential applications of data value. First, using the definition of the Data Shapley, we interpret each data source's value as an indicator of its quality and whether its presence helps or hurst the overall performance of the predictive model. We could use this information either to remove hurtful data or to identify data that has higher value and gather similar data. Secondly, \alg provides us with a simple method to for domain adaptation; that is, to adapt the performance of a given train data to a different test data set.

\subsection*{Identifying data quality}

    \alg determines the value of a data source as its approximate expected contribution to the trained model's performance if concatenated with any subset of the rest of training data sources. Therefore, data sources with high value are ones that on average contribute significantly to the trained model's performance. Similarly, low value data either do not contribute or even worsen the performance (in the case of negative value). In what follows, we apply \alg to several real world data sets and utilize the quality of data for real world applications.

    \paragraph{Value of low quality data}
        
\begin{figure}
            \centering
            \includegraphics[width=\linewidth]{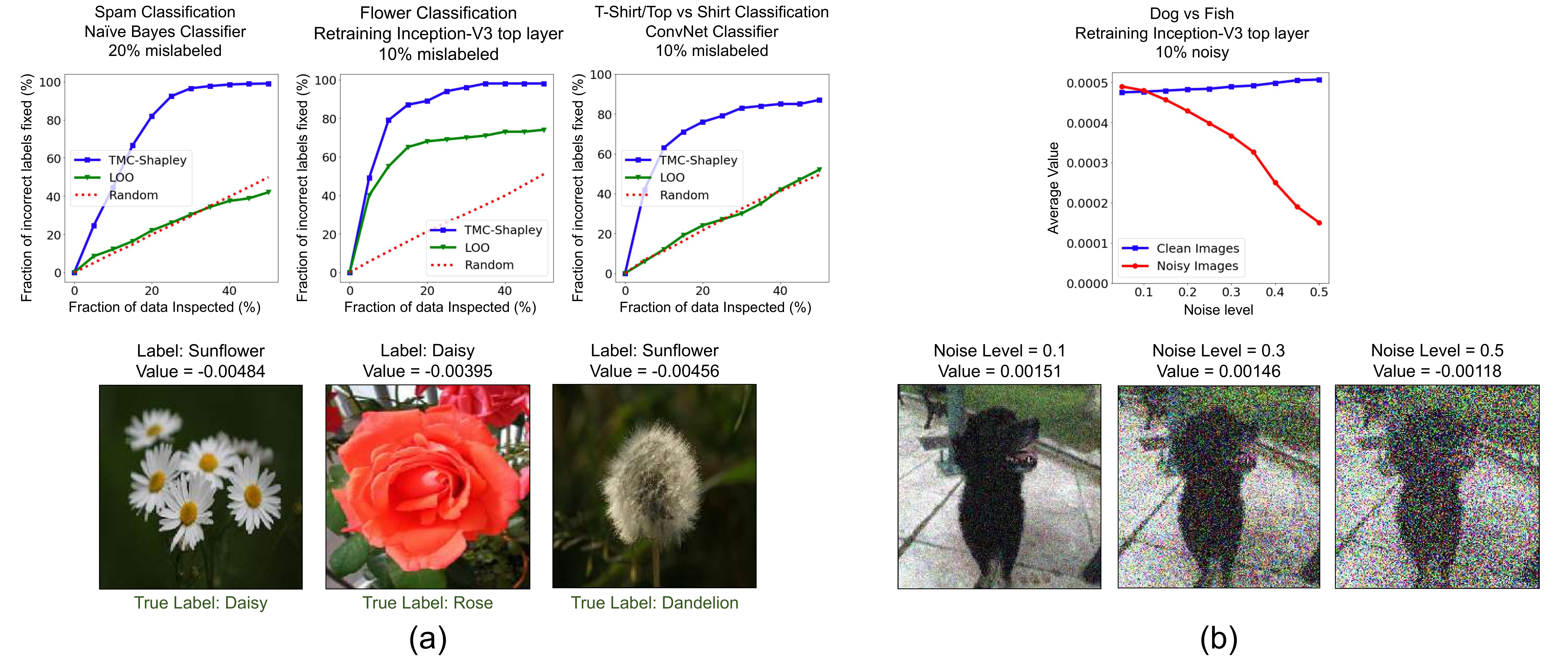} 
            \caption{(a) We inspect train data points from the least valuable to the most valuable and identify the mislabeled examples. As it is shown, by using Shapley value we need the least number of inspections for detecting mislabeled data. While leave-one-out works reasonably well on the Logistic Regression model, it's performance on the two other models is similar to random inspection. (b) We add white noise to $10\%$ of train data images. As the noise level increases, the average value of noisy images compared to clean images decreases. Each point on the plots is the average result for 10 repeats of the experiments where each time a different subset of train data is corrupted.
            \label{fig:label_flip}}
        \end{figure}

        We start by investigating the low value sources of data, how they compare to other sources and how knowing their low value could be useful. A simple real-world scenario is having a data set where some data points are mislabeled. It is known that labeling data sets is susceptible to mistakes~\cite{frenay201llabelnoise}. Additionally, mislabeling the data can be used as a simple data poisoning method~\cite{steinhardt2017poisoning}. We expected that mislabeled data points would have low Data Shapley value as they add incorrect information to the learning task and therefore are expected to harm the predictive model's prediction performance. As an experiment, we consider three data sets with mislabeled examples each used for learning a different model. Our task is to use value to find the mislabeled data points and correct the labels. For this goal, after applying \alg, we inspect the data points from the least valuable to the most valuable. As Fig.~\ref{fig:label_flip}(a) shows, in all cases, mislabeled data points are the among the ones with the least value and are discovered early on. 
        Comparing the performance of using \alg to that of leave-one-out benchmark shows that the Shapley value is a better reflection of data quality. More details of the experiment are described in Appendix~\ref{app:experiments}.
    
        In addition to the label noise, another real world scenario is when the data itself is noisy. As our second experiment, we corrupt part of the train data with noise and see how it affects value. We use Dog vs. Fish data set ~\cite{koh2017understanding} extracted from ImageNet~\cite{russakovsky2015ImageNet} and choose 500 images of each class to form the train data. The predictive model is a pre-trained Inception-v3 ImageNet classifier~\cite{szegedy2016inception} with all layers but the top layer frozen. We then corrupt $10\%$ of the train data images by adding white noise. As expected, the value of noisy images compared to the clean ones decreases if more noise is added (Fig.~\ref{fig:label_flip}(b)). 
    
    \paragraph{All data sources are not created equal}
        
        In many applications, including healthcare, data is gathered from individuals and it it is necessary to compensate them for their data. As a real-worlds example, we use \alg to compute individuals' data value in a disease prediction problem and compare low and high value data. We use the patient data in the UK Biobank data set~\cite{sudlow2015ukb} for the binary prediction task of whether an individual will be diagnosed with a certain disease in the future. We target Malignant neoplasm of breast and skin (ICD10 codes C50 and C44) as an example using 285 phenotypic features of each patient. A balanced binary data set (500 negative and 500 positive patients) is used as train data for each disease prediction task. Training Logistic regression models results in 68.7\% and 56.4\% prediction accuracy for breast and skin cancer prediction, respectively on a held out set of 1000 patients. We chose these two tasks with different levels of accuracy to investigate the outcome of \alg for tasks of various difficulty.
        
\begin{figure}
            \centering
            \includegraphics[width=\linewidth]{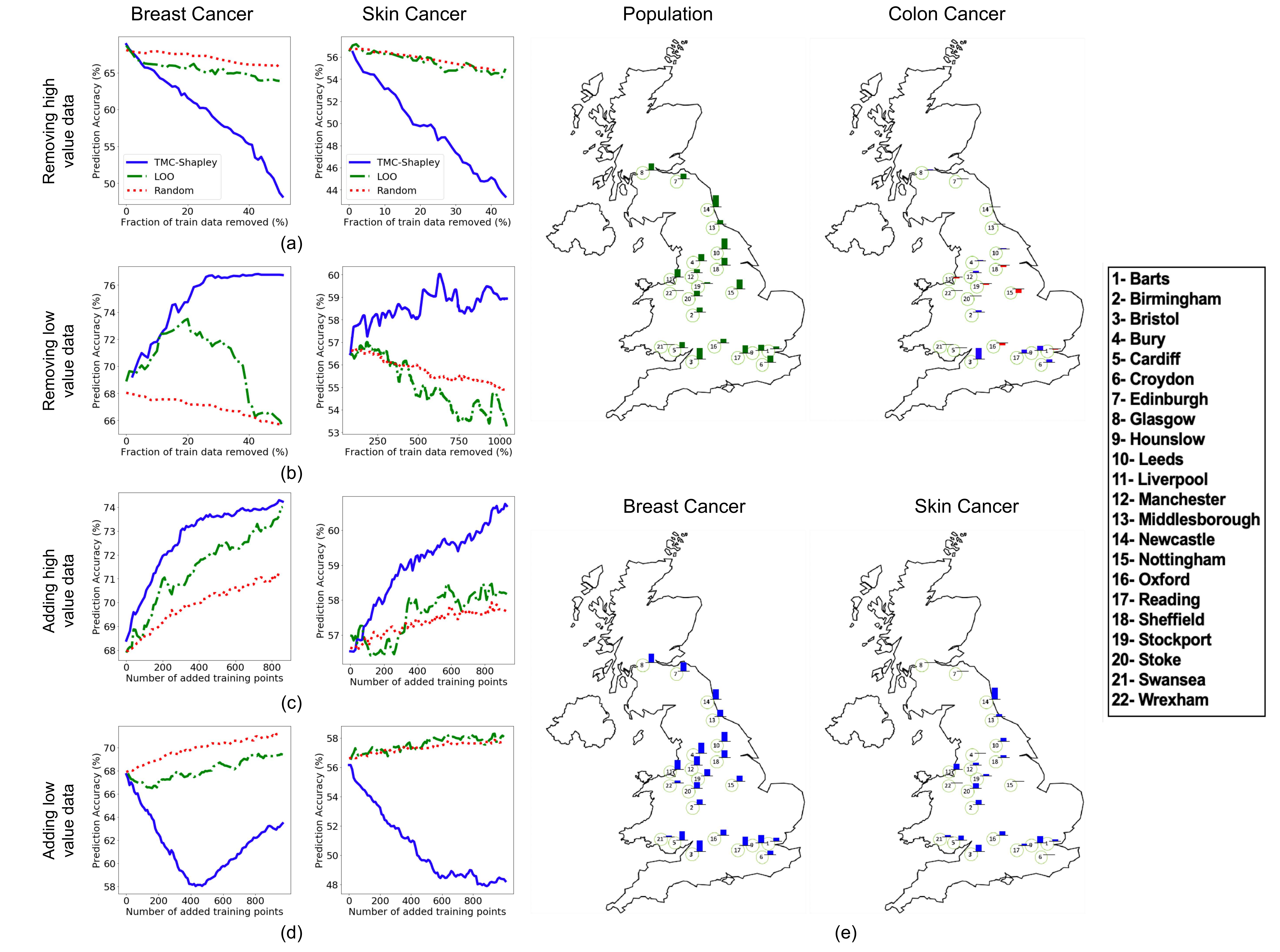} 
            \caption{\textbf{Patient Value for Disease Prediction} For breast and skin cancer prediction tasks, we  calculate the value of patient in the training data. (a) We remove the most valuable data from the train set and track the performance degradation. (b) Removing low value training data improves the predictor performance. (c) Acquiring new patients similar to high value training points improves performance more than adding patients randomly.(d) Acquiring new patients who are similar to low value points does not help. (e) Map of values of different centers across UK .
            \label{fig:ukb}}
        \end{figure}

        As our first experiment, we remove data points one by one starting with the order of their value; from the most valuable to the least valuable point. Each time, after the point is removed, a new model is trained on the remaining train data. Fig.~\ref{fig:ukb}(a) shows how the prediction performance of the trained models evolves through the experiment (accuracy on the held-out set); points that \alg considers valuable are crucially important for prediction.
        
        We can perform the experiment in reverse; Fig.~\ref{fig:ukb}(b) depicts the results for the opposite setting where we remove data points starting from the least valuable (reverse value order). It is interesting to notice that removing points with low Shapley value actually helps with better performance.
    
        In addition to removing data from the train set, we examine the opposite setting of adding data. Inspecting the train data points with high Shapley value can inform us about which new data points to collect---by recruiting similar individuals---in order to improve the model performance.
        As our second experiment, we examine a real-world scenario: we aim to add a number of new patients to the train data to improve the predictive model. We have a pool of 2000 new candidate patients to add and adding each new individual carries a cost, so we have to choose the ones that improve the performance the most. As mentioned, we want to choose patients whose data is similar to that of high value data points in train set. We fit a Random Forest regression model to the calculated values of train data points; the model learns to predict a data point's value given its observables (features and label). We then apply this regression model to the patients in the pool; in other words, we estimate the value of each patient in the patient pool by passing his or her observables through the Random Forest model. 
        
        Fig.~\ref{fig:ukb}(c) depicts how the trained model's performance evolves as we add new patients one by one by the order of their estimated value. The model's performance increases more effectively than adding new patients randomly or using LOO score as value. The opposite case is shown in Fig.~\ref{fig:ukb}(d) where adding the data points with low estimated value can actually hurt the trained model's performance.
        
        In a real-world setting, data is gathered from different sources which can be quite similar or different both in distribution and quality. Rather than valuating individual data points, one could use \alg to valuate the various sources of data. As an experiment, we use UK Biobank data set for the task of predicting future diagnosis of colon caner (ICD10 code C18), skin cancer, and breast cancer with train data of sizes 3180, 12838, and 9144 respectively. The data is gathered from 22 different health centers accross the UK and the model we use is a Random Forest binary classification model. Fig.~\ref{fig:ukb}(e) shows the relative number of patients in each center along with the Data Shapley value of each center for each of the three disease prediction tasks. It is interesting to notice that for the task of colon cancer prediction, one of the centers (14-Nottingham) is assigned with a noticeably negative value;  we further investigate why this occurs. We hypothesise that the negative value is due to the difference in the distribution of phenotypic features in that center compared to the general population. For instance, when we train the model on the general population, using the predictor's feature importance scores, age has the highest importance and then mean corpuscular haemoglobin helps with the prediction. Looking at the feature importance scores of the model trained on the entire population, age is the single strongest feature for predicting colon cancer---older age increases the chance of diagnosis (KS test, $p = 1.5\mathrm{e}{-6}$). However, in Nottingham, there is no significant distributional difference between the age of healthy and diagnosed patients (KS test, $p = 0.14 $).

\subsection*{Using value to adapt to new data}

\begin{figure}
        \centering
        \includegraphics[width=\linewidth]{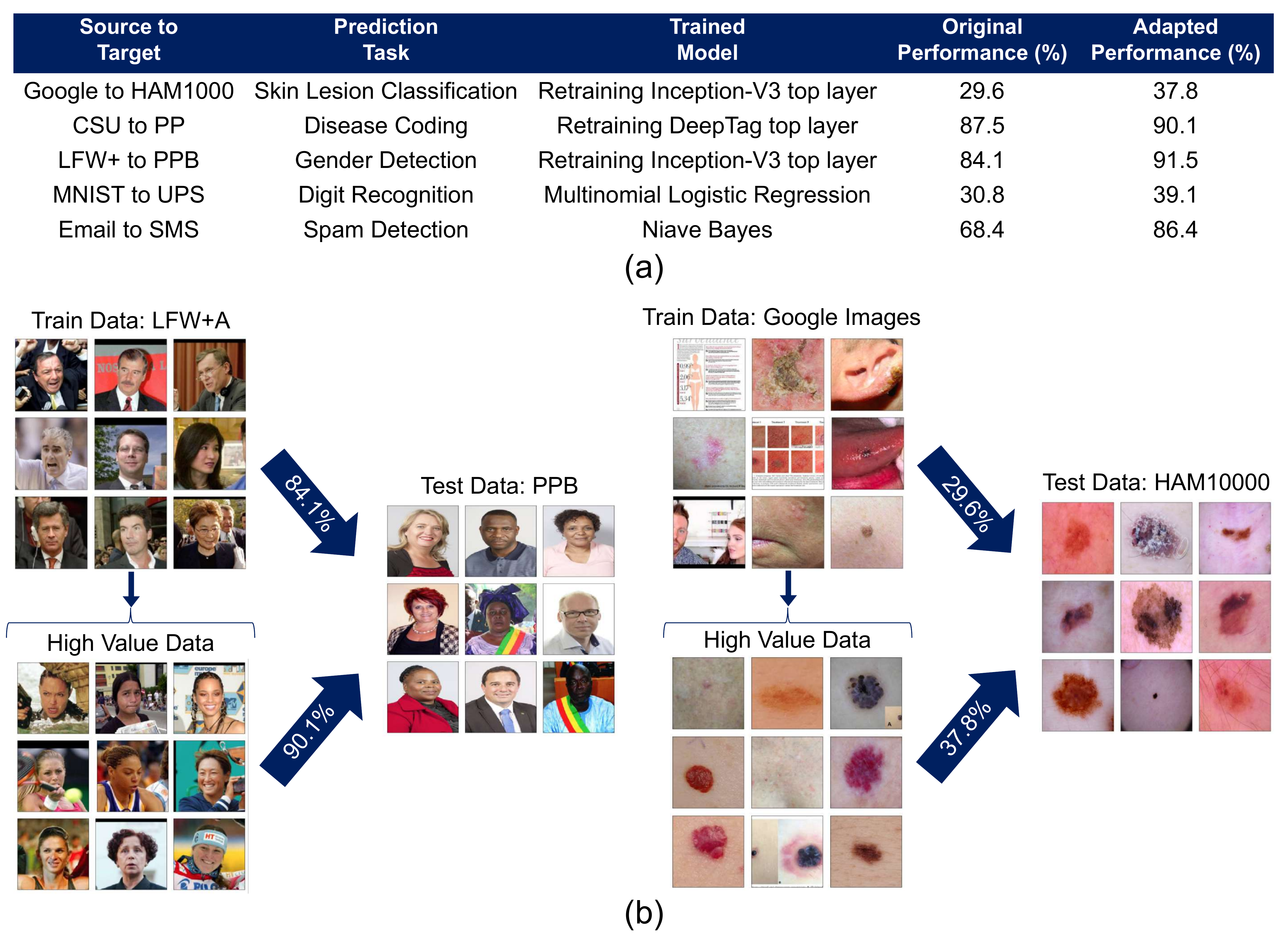} 
        \caption{{Data shapley value for domain adaptation} Adapting to a new data set. Available training data is not always completely similar to the test data. By valuating the training set data points, we can first, remove points with negative value and then, emphasize the importance of valuable points by assigning more weight during training.
        \label{fig:adaptation}}
    \end{figure}

    In real-world settings, the data used for training the machine learning model could be different from the data the model will be deployed and tested on~\cite{torralba2011unbiased}. For example, settings where gathering data similar to the deployment data (target) we care about is expensive and we only have access to train data (source) that is different from the test statistically or quality-wise~\cite{gebru2017fine}. In these settings, the question related to data valuation is whether there are data sources in the training data that help the most with the task of adaptation or on the other side, whether there are train sources that harm the adaptation task. We use \alg to compute the value of the train data sources similar to previous experiments with the only difference that the trained model's performance score is now evaluated on the target data. After valuating the source data points, first, we remove points with negative value. Secondly, we use Data Shapley value to emphasize the effect of points that help the most with the adaptation task. We train a new model using a weighted loss function where each point's weight in the new loss function is its relative value. Fig.~\ref{fig:adaptation} shows the results on a held-out set of data points from the target data (data points separate from ones used in \alg for evaluating model's performance).

    \paragraph{Skin pigmented lesion detection}
        As a real-world scenario, we examine the case of adapting from cheap low quality data to high quality data created by field experts. The prediction task is to classify $7$ different family of skin lesions. Creating a high quality data set of skin lesions requires field experts to take the lesion images and label each image. Instead, we created a cheap train data by searching each lesions name (with simple added keywords) using google image search tool. The target data, however, is created and labeled by field experts. HAM10000\cite{tschandl2018ham10000} data set contains dermoscopic images of pigmented lesions where each image is taken using a dermatoscope by field experts. We train the model using our cheap train data that contains 1155 images (165 image of each lesion). We expect our train data to contain mislabeled, unrelated, low-quality, and  non-dermatoscopic images (Fig.~\ref{fig:adaptation}(b)). The model performance is evaluated on a balanced subset of 400 HAM1000 images. We train a pre-trained Inception-V3 ImageNet classifier with all layers frozen but the top layer. After computing the values of train data points, we delete points with negative Shapley value and reweight the remaining points with their relative positive value and retrain the model. The prediction accuracy of the model increases from 29.6\% accuracy to 37.8\% on a balanced held-out set of 400 data points from HAM1000 data set.

    \paragraph{Fairness in gender detection}
    
        Buolamwini \& Gebru~\cite{buolamwini2018gender}  discussed that machine learning models for the task of human face gender detection have degraded performances when it comes to minorities and women. Following the experiment setting discussed by Kim et al. \cite{kim2018multiaccuracy}, we train an Inception-Resnet-V1\cite{szegedy2017inception_resnet} model for gender classification using the Celeb A\cite{liu2015celeba} data set with more than 200,000 face images. We then freeze all layers of the network except the top layer. The adaptation task is to train the model on LFW+A\cite{wolf2011lfw1} data set and test it on the PPB\cite{buolamwini2018gender} data set. LFW+A data set has unequal representation of minorities and women ($21\%$ female, $5\%$ black) while PPB data set is designed to have equal representation of sex and skin color intersections. Using the introduced adaptation scheme, the gender detection accuracy increases from $84.1\%$ to $91.5\%$ on a held-out set of 400 PPB images. It is insightful to mention that all of the LFW+A images with negative value are from male subjects (the over-represented groups) while the top $20\%$ most valuable images are of the female subjects (Fig.~\ref{fig:adaptation}(b)).

    \paragraph{Difference in clinical notes}
        
        Most veterinary visits are recorded in free-text notes.
        Nie et. al.\cite{nie2018deeptag} introduced DeepTag, a machine learning based method, to automatically tag visit notes with the relevant disease codes. The issue, however, is that different institutions can have different writing styles. They trained an LSTM model on more than 100000 clinical visits from Colorodo State University of Veterinary Medicine and Biomedical Sciences (CSU data set). We we freeze all but the last layer of their model. In our adaptation task, the target data is a set of 5000 clinical notes from the same CSU data set separate from the data used for training the LSTM model. The target data is a set of 400 clinical notes from a Private Practice (PP data set) that is different in different writing style and institution type from the source data. Average prediction accuracy increases from $87.5\%$ to $90.1\%$ for detecting the 10 most frequent diseases.
        
    \paragraph{Further Experiments} 
        We performed two more domain adaptation experiment. First, we train a multinomial logistic regression classifier on 1000 handwritten digits of MNIST data set~\cite{lecun1998mnist} while the target data set is the USPS digits ~\cite{hull1994database} data. As our second experiment, we train a Naive Bayes model on an Email spam detection data set~\cite{metsis2006spam} and test it on the target data set of SMS spam  detection~\cite{almeida2013towards}. The prediction accuracy on held-out sets of target data improves in both settings.

\section*{Related Works}

Shapley value was proposed in a classic paper in game theory \cite{shapley1953value} and has been widely influential in economics \cite{shapley1988shapley}. It has been applied to analyze and model diverse problems including voting, resource allocation and bargaining \cite{milnor1978values, gul1989bargaining}. To the best of our knowledge, Shapley value has not been used to quantify  data value in a machine learning context like ours. Shapley value has been recently proposed as a feature importance score for the purpose of interpreting black-box predictive models \cite{kononenko2010shap1,datta2016shap2,lundberg2017shap3, cohen2007shap4, chen2018shapley, lundberg2018consistent}. Their goal  is to quantify, for a given prediction, which features are the most influential for the model output. Our goal is very different in that we aim to quantify the value of individual data points (not features). There is also a literature in estimating Shapley value using Monte Carlo methods, network approximations, as well as analytically solving Shapley value in specialized settings \cite{fatima2008linear, michalak2013efficient, castro2009approx1, maleki2013bounding, hamers2016new} Parallel works have studied Shapley value in the context of data valuation focusing on approximation methods and applications in a data market.~\cite{jia2019towards, agarwal2018marketplace}.

In linear regression, Cook's Distance measures the effect of deleting one point on the regression model \cite{cook1977detection}. Leverage and influence are related notions that measures how perturbing each point affects the model parameters and model predictions on other data \cite{cook1982residuals, koh2017understanding}. These methods, however, do not satisfy any equitability conditions, and also have been shown to have robustness issues \cite{ghorbani2017interpretation}.
In the broad discourse, value of data and how individuals should be compensated has been intensely discussed by economists and policy makers  along with the discussion of incentivizing participants to generate useful data.\cite{arrieta2017should, posner2018radical}

\section*{Discussion}

    We proposed \alg as an equitable framework to quantify the value of individual training sources. \alg uniquely satisfies three natural properties of equitable data valuation. There are ML settings where these properties may not be desirable and perhaps other properties need to be added. It is a very important direction of future work to clearly understand these different scenarios and study the appropriate notions of data value. Drawing on the connections from economics, we believe the three properties we listed is a reasonable starting point. 
    While our experiments demonstrate several desirable features of \alg, we should interpret it with care. Due to the space limit, we have skipped over many important considerations about the intrinsic value of personal data, and we focused on valuation in the very specific context of training set for supervised learning algorithms. We acknowledge that there are nuances in the value of data---e.g. privacy, personal association---not captured by our framework. Moreover we do not propose that people should be exactly compensated by their value; we believe \alg is more useful for the quantitative insight it provides.  
    
    In the \alg framework, the value of individual datum depends on the learning algorithm, evaluation metric as well as other data points in the training set. Therefore when we discuss data with high (or low) Shapley value, all of this context is assumed to be given. A datum that is not valuable for one context could be very valuable if the context changes. Understanding how \alg behaves for different learning functions and metrics is an interesting direction of follow up work.



\pagebreak
\bibliographystyle{naturemag}
\bibliography{sample}


\begin{itemize}

\item Python code implementation is available at

\url{https://github.com/amiratag/DataShapley}.

\item A.G. is supported by a Stanford Graduate Fellowship. J.Z. is supported by NSF CRII 1657155, 1RM1HG010023-01, a Chan-Zuckerberg Investigatorship and grants from the Silicon Valley Foundation.

\item The authors declare that they have no
competing financial interests.

\item Correspondence and requests for materials
should be addressed to J.Z.~(email: jamesz@stanford.edu).

\end{itemize}
 
\newpage
\appendix
\section{TMC-Shapley algorithm}
\label{app:TMC-Shapley}
    \subsection*{Approximating Shapley value of data sources}
    
        The Shapley formula in Eqn.~\ref{eq:shapley} uniquely provides  an equitable assignment of values to data points.
        Computing shapley values, however, requires computing all the possible marginal contributions which is exponentially large in the train data size. In addition, for each $S \subseteq D$, computing $V(S)$ involves learning a predictor on $S$ using the learning algorithm $\mathcal{A}$. As a consequence, calculating the exact Shapley value is not tractable for real world data sets. In this section, we discuss approximation methods to estimate the  value.
    
        \paragraph{Monte-Carlo method: }
        
             We can rewrite Eqn.~\ref{eq:shapley} into an equivalent formulation by setting $C = 1/n!$. Let $\Pi$ be the uniform distribution over all $n!$ permutations of data points, we have:
            $$
                \phi_i = \mathbb{E}_{\pi \sim \Pi} [V(S_\pi^i \cup \{i\}) - V(S_\pi^i)]$$
            
            where $S_{\pi}^{i}$ is the set of data points coming before datum $i$ in permutation $\pi$ ($S_{\pi}^{i}= \emptyset$ if $i$ is the first element).
            
            As described in Eqn.~\ref{eq:shapley}, calculating the Shapley value can be represented as an expectation calculation problem. Therefore, Monte-Carlo method have been developed and analyzed to estimate the Shapley value ~\cite{mann196large, castro2009polynomial, maleki2013bounding}. First, we sample a random permutations of data points. Then, we scan the permutation from the first element to the last element and calculate the marginal contribution of every new data point. Repeating the same procedure over multiple Monte Carlo permutations, the final estimation of the Shapley values is simply the average of all the calculated marginal contributions. This Monte Carlo sampling gives an unbiased estimate of the Shapley values. In practice, we generate Monte Carlo estimates until the average has empirically converged. Previous work has analyzed error bounds of Monte-carlo approximation of Shapley value \cite{maleki2013bounding}. 
            
        \paragraph{Truncation: } 
            In the machine learning setting, $V(S)$ for $S \subseteq N$ is usually the predictive performance of the model learned using $S$ on a separate test set. Because the test set  is finite, $V(S)$ is itself an approximation to the true performance of the trained model on the test distribution, which we do not know. In practice, it is sufficient to estimate the Shapley value up to the intrinsic noise in $V(S)$, which can be quantified by measuring variation in the performance of the same predictor across bootstrap samples of the test set \cite{friedman2001elements}.  
            On the other hand, as the size of $S$ increases, the change in performance by adding only one more training point becomes smaller and smaller ~\cite{mahajan2018exploring, beleites2013sample}. Combining these two observations lead to a natural truncation approach. 
    
            We can define a ``performance tolerance'' based on the bootstrap variation in $V$. As we scan through a sampled permutation and calculate marginal contributions, we truncate the calculation of marginal contributions in a sampled permutation whenever $V(S)$ is within the performance tolerance of $V(D)$ and set the marginal contribution to be zero for the rest of data points in this permutation. Appendix~\ref{app:approx} shows that truncation leads to substantial computational savings without introducing significant estimation bias.
            In the rest of the paper, we refer to the combination of truncation with Monte-Carlo as the  ``Trunctated Monte Carlo Shapley''(TMC-Shapley); described with more details in Algorithm~\ref{alg:TMC-Shapley}.
\section{G-Shapley algorithm}
\label{app:G-shapley}

    For a wide family of predictive models, $\mathcal{A}$ involves a variation of stochastic gradient descent where randomly selected batches of $D$ update the model parameters iteratively. One simple approximation of a completely trained model in these settings is to consider training the model with only one pass through the training data; in other words, we train the model for one ``epoch'' of $D$. This approximation fits nicely within the framework of Algorithm~\ref{alg:TMC-Shapley}: for a sampled permutation of data points, update the model by performing gradient descent on one data point at a time; the marginal contribution is the change in model's performance. Details are described in Algorithm ~\ref{alg:G-Shapley}, which we call Gradient Shapley or G-Shapley for short. In order to have the best approximation, we perform hyper-parameter search for the learning algorithm to find the one resulting best performance for a model trained on only one pass of the data which, in our experiments, result in learning rates bigger than ones used for multi-epoch model training.  Appendix~\ref{app:approx} discusses numerical examples of how good of an approximation G-Shapley method yields in this work's experimental results.
    \begin{algorithm}[tb]
      \caption{\textbf{Gradient Shapley}}
      \label{alg:G-Shapley}
          \begin{algorithmic}
            \STATE {\bfseries Input:} Parametrized and differentiable loss function $\mathscr{L}(.; \theta)$, train data $D = \{1,\dots,n\}$ ,  performance score function $V(\theta)$
            \STATE {\bfseries Output:} Shapley value of training points: $\phi_1,\dots,\phi_n$
            \vspace{1mm}
            \STATE Initialize $\phi_i=0$ for $i=1,\dots,n$ and $t=0$
            \WHILE{Convergence criteria not met}
                \STATE $t \leftarrow t+1$
                \STATE $\pi^{t}$: Random permutation of train data points
                \STATE $\theta^t_0 \leftarrow \mbox{Random parameters}$
                \STATE $v^t_0 \leftarrow V(\theta^t_0)$ 
                \FOR{$j \in \{1,\ldots,n\}$}
                    \vspace{0.5mm}
                    \STATE $\theta^{t}_j \leftarrow \theta^{t}_{j-1} - \alpha  \nabla_{\theta} \mathscr{L}(\pi^t[j];\theta_{j-1})$
                    \vspace{0.5mm}
                    \STATE $v^t_j \leftarrow V(\theta^{t}_j)$ 
                    \STATE $\phi_{\pi^t[j]} \leftarrow \frac{t-1}{t}\phi_{\pi^{t-1}[j]} + \frac{1}{t}(v^t_{j} - v^t_{j-1}) $
                    \vspace{1mm}
                 \ENDFOR
            \ENDWHILE
            \vspace{1mm}
          \end{algorithmic}
    \end{algorithm}
    
    We repeat the Label-Flip detection  ( Fig.~\ref{fig:label_flip}(a)) and patient data valuation experiments(Fig.~\ref{fig:ukb}(a-d)) experiments using the G-Shapley valuation method. Supp. Fig.~\ref{fig:G-Shapley} describes the results. As it is shown, the results using G-Shapley approximation method are similar to using the TMC-Shapley method and better than the Leave-One-Out benchmark.
    
\setcounter{figure}{0}    
\renewcommand{\figurename}{Supplementary Figure}
    
        \begin{figure*}[ht]
            \centering
            \includegraphics[width=\linewidth]{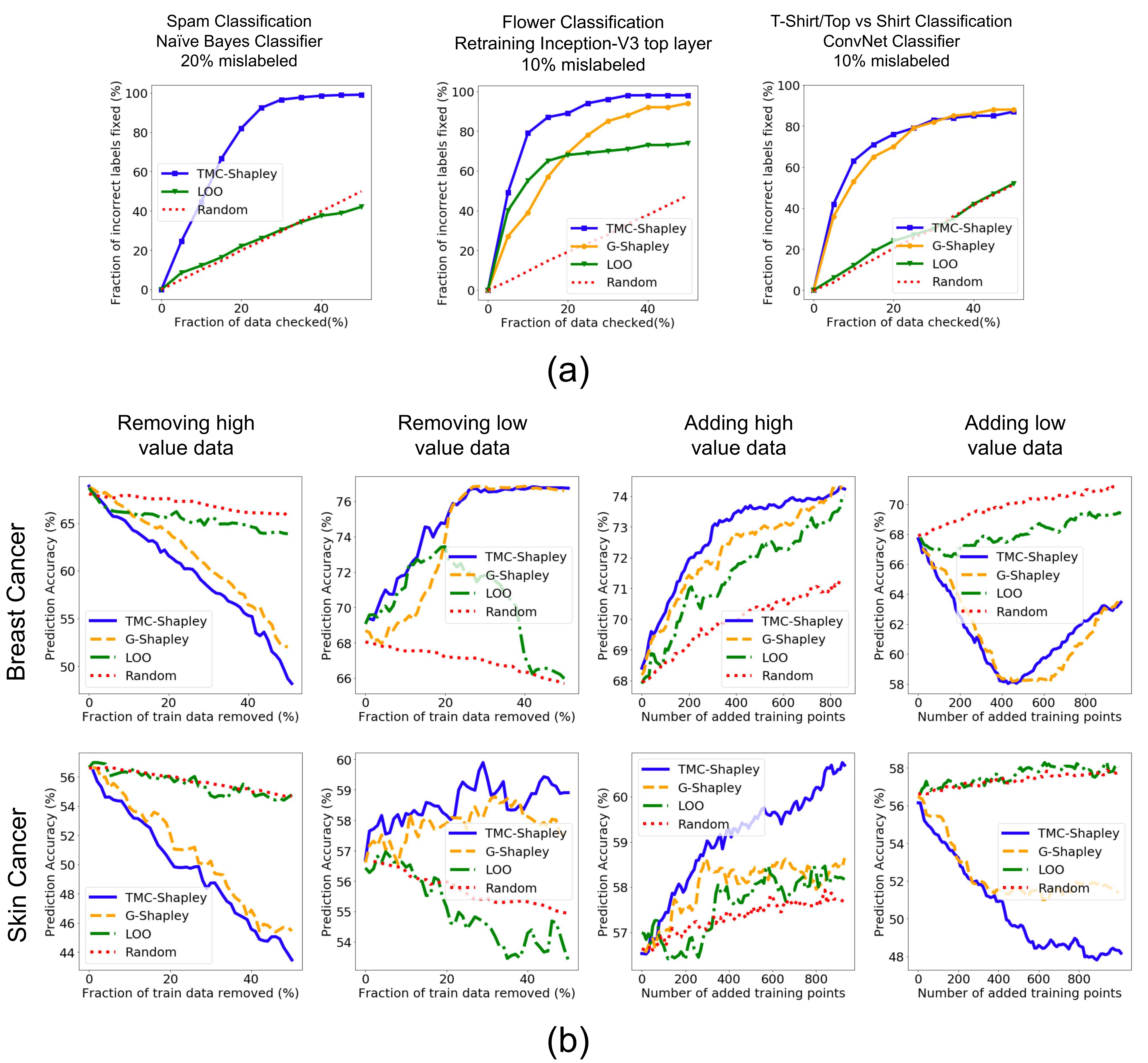} 
            \caption{\textbf{G-Shapley algorithm}\label{fig:G-Shapley}}
        \end{figure*}
        
\section{Experimental Details}
\label{app:experiments}

    \paragraph{Value of low quality data}
        The experiment in Fig.~\ref{fig:label_flip}(a) was performed on three different data sets and three different predictive models: 1- A Multinomial Naive Bayes model trained on 3000 training points of the the spam classification data set~\cite{metsis2006spam}  where 600 points have flipped labels. 2- Multinomial Logistic Regression model trained on 1000 images from flower classification data set\footnote{adapted from https://goo.gl/Xgr1a1} (5 different classes) where 100 images are mislabeled. We pass the flower images through Inception-V3 model and train the multinomial logistic regression model on the Inception-V3's representation. 3- A convolutional neural network with one convolutional and two feed-forward layers trained on 1000 images from the Fashion MNIST data set\cite{xiao2017fashion} to classify T-Shirts and Tops versus Shirts. 100 images are mislabeled. In all cases, the value is computed on a separate set of size 1000 data points.
        
        The experiment in Fig.~\ref{fig:label_flip}(b) was performed on the Dog vs. Fish data set~\cite{koh2017understanding} which consists of 1200 images of different dogs and 1200 images of fish are samples from the original ImageNet~\cite{russakovsky2015ImageNet} data set. We randomly select 500 images of each class as our training set and pass the images through the Inception-V3 network to get the image representation of the network. A logistic regression model is then learned on top of the Inception-V3 representation. We fix a noise level and choose a random set of 100 images and corrupt them with gaussian noise. For pixels normalized to have values between zero and one, noise-level stands for the standard deviation of the added guassian noise. We then calculate the Shapley values. The same experiment is repeated 10 times with 10 different sets of noisy images for the fixed noise level. We then used other noise levels and reported the average results for each noise level.
        
    \paragraph{All data points are not created equal}
    
        In this experiment, for each disease, from the original data set of  500,000 patients, we choose all the patients that were diagnosed with the disease within two years of their visit. To create a balanced binary data set of disease prediction task, we then sample the same number of patients that were not diagnosed. For the experiments in Fig.~\ref{fig:ukb}(a-d), we sample a random set of 500 patients of each class. We then train a logistic regression model and use a separate set of 500 patients of each class as our test set. We use the prediction accuracy of the trained model on this test set as our evaluation metric to compute the Shapley values. To prevent the possibility of indirect over-fitting to test set, to report the results in the figures, a separate held-out set of 500 patients of each class used in the plots of Fig.~\ref{fig:ukb}(a-d). 
        
        For experiments in Fig.~\ref{fig:ukb}(e-g) the same method is used to create the data set but instead of computing values for each patient's data, we compute values for each center. For each disease, $80\%$ of the original data set of that disease is used as training set and $10\%$ of the remaining data set is used as the test set for computation of evaluation metric.

\section{Additional Experiments}
\label{app:additional}

    \paragraph{Synthetic Experiments}

\begin{figure*}[ht]
            \centering
            \includegraphics[width= 0.7\linewidth]{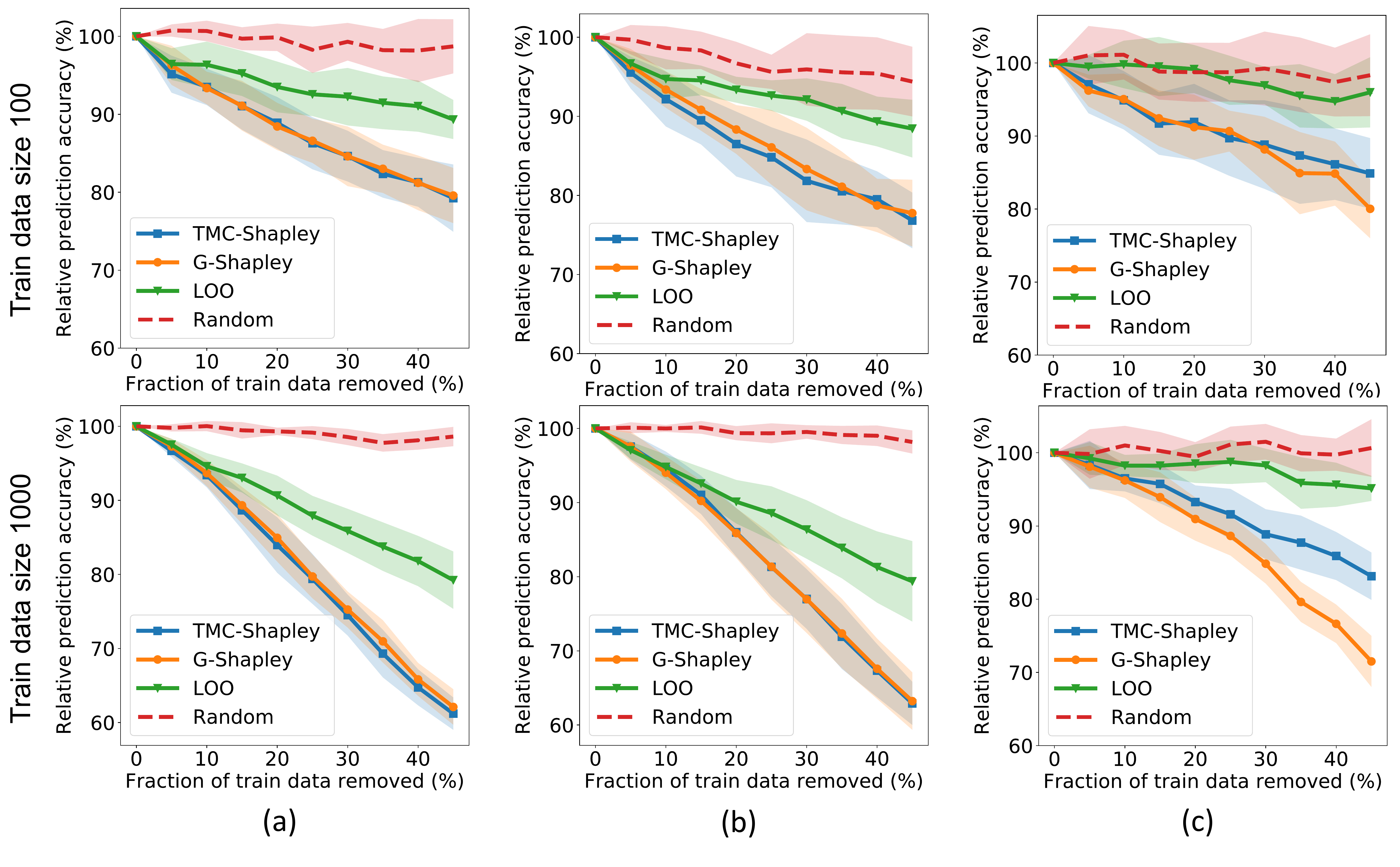} 
            \caption{\textbf{Synthetic experiments} Average results are displayed for three different settings. Vertical axis if relative accuracy which stands for accuracy divided by the accuracy of the model trained on the whole train data without any removal. For each figure, 20 data sets are used. In all data sets, the generative process is as follows: for input features $\mathbf{x}$, the label is generated such that $p(y|x)=f(x)$ where in (a) $f(.)$ is linear and in (b) $f(.)$ is a third order polynomial and (c) uses the same data sets as (b). In (a) and (b) the model is logistic regression and in (c) it's a neural network.  Both Shapley methods do a better job at assigning high value to data points with highest positive effect on model performance. Colored shaded areas stand for standard deviation over results of 20 data sets.
            \label{fig:synthetic}
            }
        \end{figure*}

        We use synthetic data to further analyze Shapley values. 
        The data generating process is as follows. First, features are sampled from a 50-dimensional  Gaussian distribution $\mathcal{N}(0, I)$ . Each sample $\mathbf{x}_i$'s label is then assigned a binary label $y_i$ where $P(y_i=1) = f(\mathbf{x})$ for a function $f(.)$. We create to sets of data sets: 20 data sets were feature-label relationship is linear (linear $f(.)$) , and 20 data sets where $f(.)$ is a third order polynomial. For the first sets of data set we us a logistic regression model and for the second set we use both a logistic regression and a neural network with one hidden layer. We then start removing training points from the most valuable to the least valuable and track the change in model performance. Supp. Fig.~\ref{fig:synthetic} shows the results for using train data size of 100 and 1000; for all of the settings, the Shapley valuation methods do a better job than the leave-one-out in determining datum with the most positive effect on model performance. Note here that Shapley value is always dependent on the chosen model: in a dataset with non-linear feature-label relationship, data points that will improve a non-linear model's performance, can be harmful to a linear model and therefore valueless. 

    \paragraph{Consistensy of value across different models}
        For the 22 centers of the UK Biobank data set mentioned in Fig.~\ref{fig:ukb}, we create balanced data sets of 10 binary prediction tasks of 10 different cancers with icd10 codes: C44, C50, C61, C18, C34, C78, C79, C67, C43, C20. For each data set, we then train three binary prediction models: Logistic Regression, Random Forest Classifier, K Nearest Neighbors Classifier and compute the data points' values for each model and disease. The average Spearmanr's rank order correlation of center values between Logistic Regression and Random forest models is equal to $50.52$ with a max of $0.80$ and a min of $0.18$. The average rank order correlation between the Logistic Regression model and the KNN model is equal to $0.32$ with a min of $-0.17$ and a max of $0.53$. For the Random Forest model and the KNN model, the average rank correlation is equal to  $0.42$ with a min of $0.16$ and a max of $0.57$. The detailed results are shown in Table.~\ref{table:versus}.
        \begin{table}
        \centering
            \begin{tabular}{c|c|c|c|c|c|c|c|c|c|c}
                 Models &  C44 & C50 & C61 & C18 & C34 & C78 & C79 & C67 & C43 & C20\\
                 \hline
                 \hline
                 \textbf{Logistic Regression} vs \textbf{Random Forest} & 0.38 & 0.65 & 0.73 & 0.47 & 0.18 & 0.46 & 0.35 & 0.72 & 0.80 & 0.44\\
                 \textbf{Logistic Regression} vs \textbf{KNN} & 0.18 & 0.53 & 0.48 & 0.33 & 0.34 & -0.17 & 0.24 & 0.52 & 0.37 & 0.35\\
                 \textbf{Random Forest} vs \textbf{KNN} & 0.57 & 0.50 & 0.33 & 0.49 & 0.45 & 0.23 & 0.53 & 0.43 & 0.53 & 0.16
            \end{tabular}
             \caption{\textbf{Value across different models} For each disease, we compute the value of 22 centers for three different models. The Spearman's rank order correlation between values of three models are shown in this table. \label{table:versus}}
        \end{table}

    \paragraph{Value of different subgroups}
        In many settings, in order to have more robust interpretations or because the training set is very large, we prefer to compute the value for groups of data points rather than for individual data. 
        For example, in a heart disease prediction setting, we could group the patients into discrete bins based on age, gender, ethnicity and other features, and then quantify the value of each bin.  In these settings, we can calculate the Shapley value of a group using the same procedure as Algorithm~\ref{alg:TMC-Shapley}, replacing the data point $i$ by group $i$. As a consequence, even for a very large data set, we can calculate the group Shapley value if the number of groups is reasonable.
        
\begin{figure}[ht]
            \centering
            \includegraphics[width=0.5\linewidth]{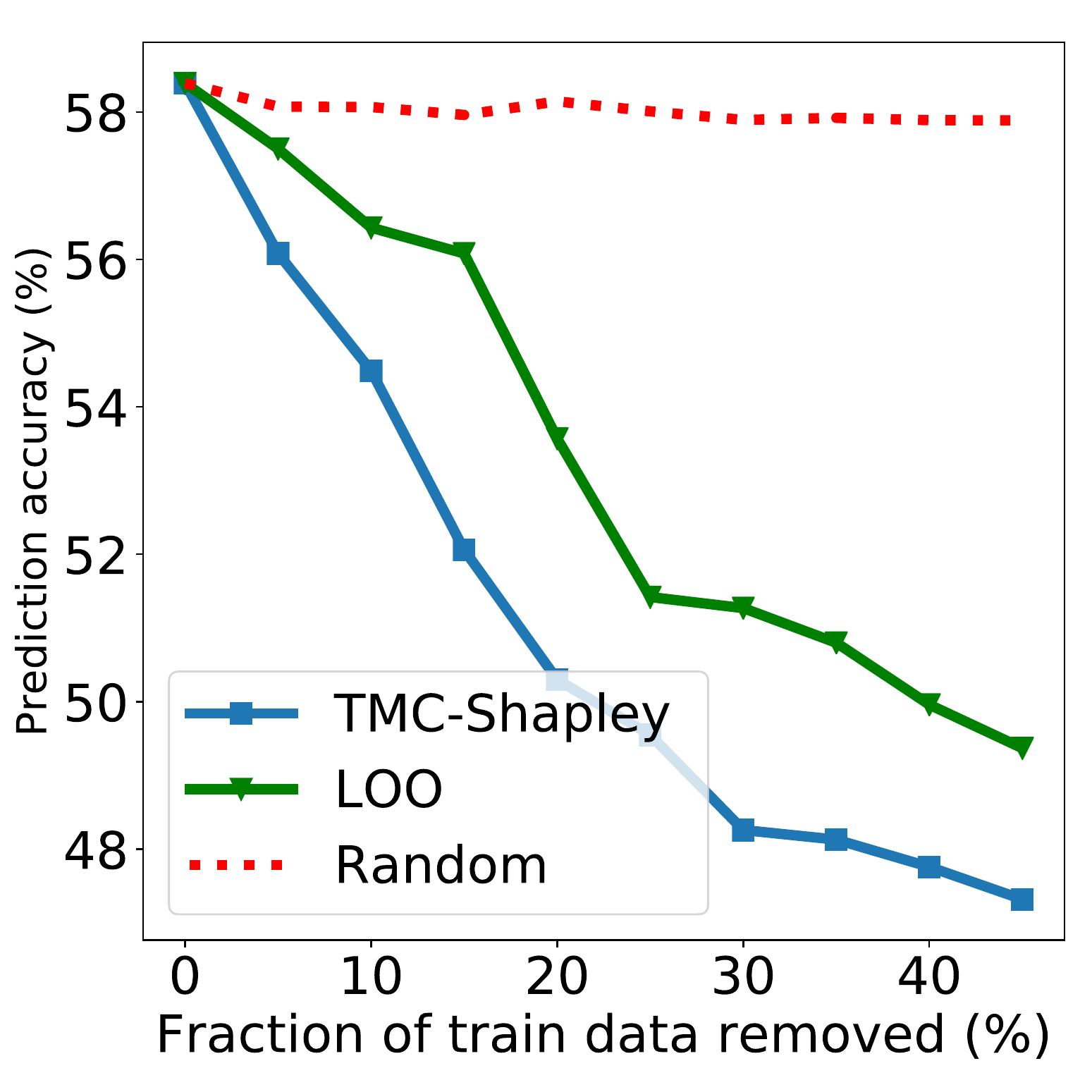}
            \caption{\textbf{Group Shapley:} We comute Shapley values for 146 different demographic groups. As shown in the plot, similar to the case of assigning value to individual data points where the high value data points are crucial for the predictive model's performance, removing more valuable groups also has a crucial impact on degrading the trained model.
            \label{fig:combined}}
        \end{figure}

        In this experiment, we use a balanced subset of the hospital readmission data set~\cite{strack2014impact} for binary prediction of a patient's readmission. We group patients into 146 groups by intersections of demographic features of gender, race, and age. A gradient boosting classifier trained on a train set of size 60000 yields and accuracy of 58.4\%. We then calculate the TMC-Shapey values of groups. Fig~\ref{fig:combined} shows that the most valuable groups are also the most important ones for model's performance. In addition to computational efficiency, an important advantage of group Shapley is its easy interpretations. For instance, in this data set, groups of older patients have higher value than younger ones, racial minorities get less value, and groups of females tend to be more valuable than males with respect to \alg, and so forth. 

\section{How good are the approximations?}
\label{app:approx}
\begin{figure*}[ht]
            \centering
            \includegraphics[width=\linewidth]{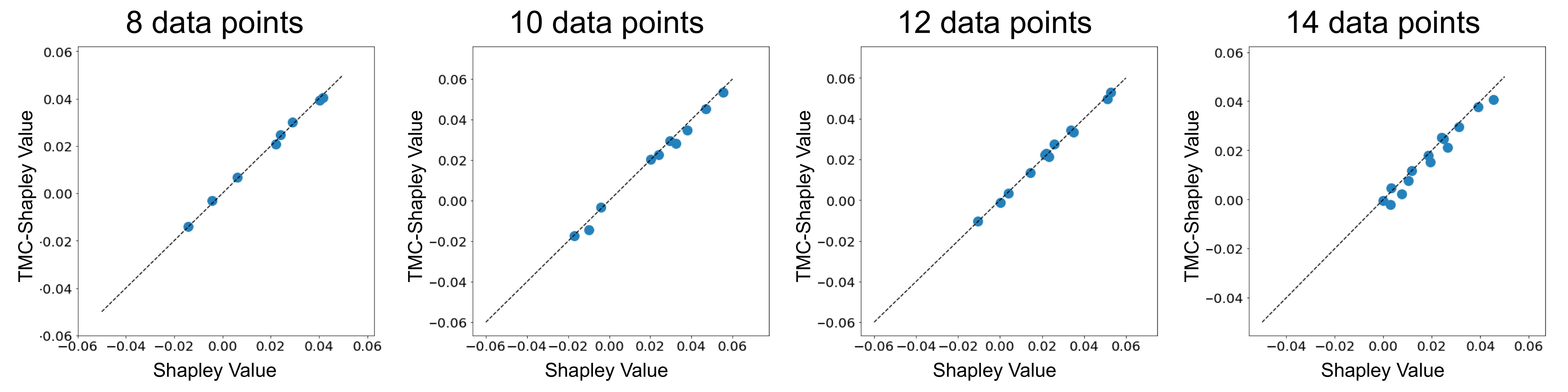} 
            \caption{\textbf{Approximation vs true value} Four examples of synthetic data sets with their respective true Shapley value and approximated TMC-Shapley value are depicted. As it is shown, the approximation manages to capture the ordering perfectly and also returns similar values.
            \label{fig:true}}
        \end{figure*}

    \paragraph{TMC-Shapley vs true Shapley value}
        We computed the true Shapley value for synthetic training data with sizes from 4 to 14 data points trained with a logistic regression model. The pearson correlation between the true Shapley value and the approximate TMC-Shapley value is in the range of $98.4\%$ to $99.5\%$. Supp. Fig.~\ref{fig:true} depicts examples.
    \paragraph{Robustness to truncation}

\begin{figure*}[ht]
            \centering
            \includegraphics[width=0.65\linewidth]{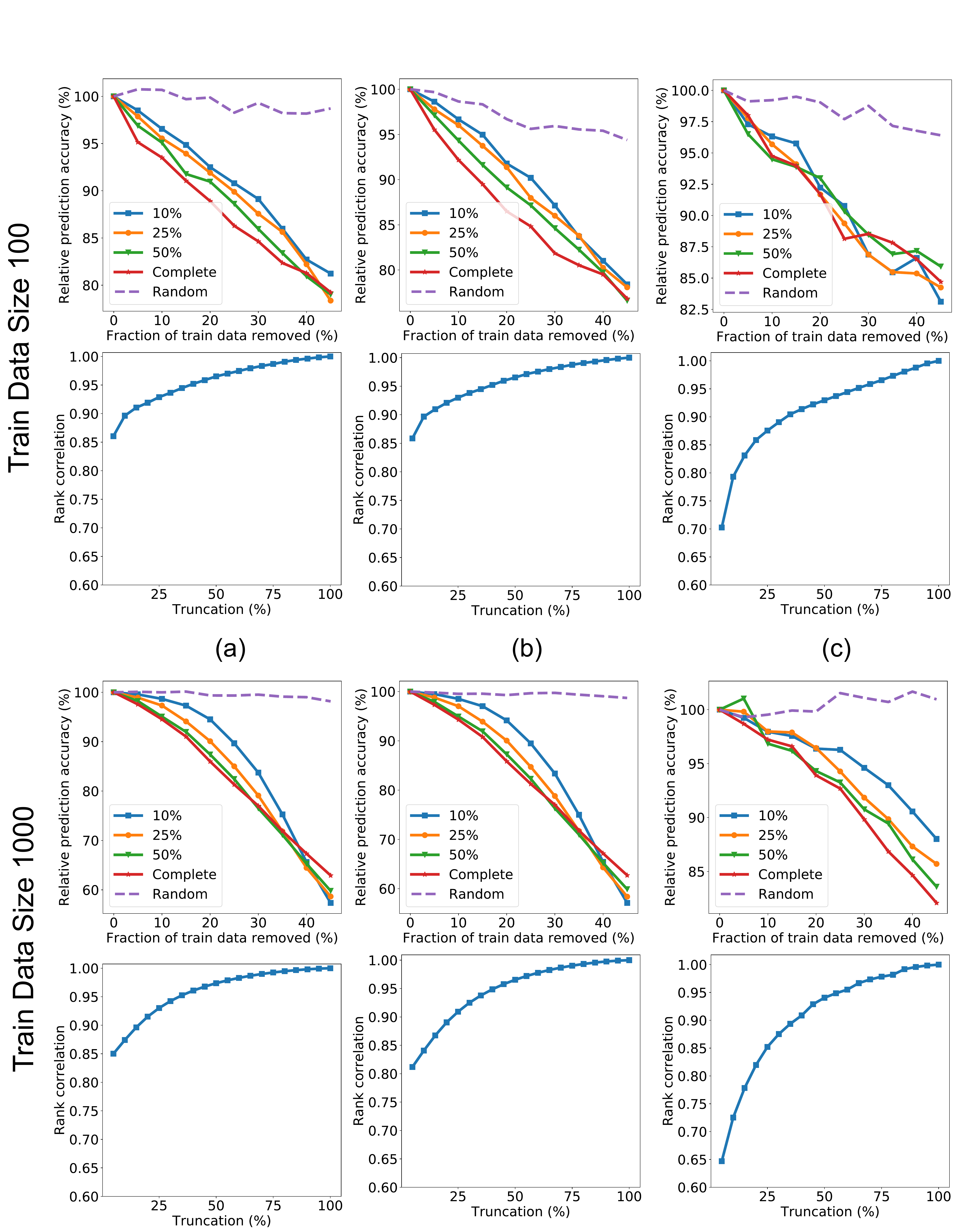} 
            \caption{\textbf{Truncation approximation} A truncation of $10\%$ means that in Alg.~\ref{alg:TMC-Shapley}, for each sampled permutations, we calculate the marginal contributions of the first 10$\%$ elements of that permutation and approximate the remaining by zero marginal contribution. For the datasets generated in Section 4.2. Columns (a), (b), and (c) use the same data sets as their corresponding column in Supp. Fig.~\ref{fig:synthetic}. For each train data size and data set, the upper plot shows the change in model performance as we remove points with high TMC-Shapley values derived by various truncation levels. The bottom plot shows the rank order correlation between TMC-Shapley values derived by a speicif truncation level and the values derived by having no truncation. It can be observed that truncation approximation's bias can be negligible.
            \label{fig:truncation}}
        \end{figure*}

        For the same data sets in Section 4.2, we do the following experiment: For each iteration of TMC-Shapley, we truncate the computation of marginal contributions at different positions. For instance, a truncation of size $10\%$ means that for a data set of size 1000, in each iteration of Alg.~\ref{alg:TMC-Shapley}, for the sampled permutation of data points we only perform the calculation of marginal contributions for the first 100 elements and approximate the rest with zero. Supp. Fig.~\ref{fig:truncation} shows the results. Columns (a), (b), (c), are from the same data sets of the corresponding columns in Supp. Fig.~\ref{fig:synthetic} and are trained using the same models. For each data set, model, and training set size, we show two plots. The first plot shows how for different levels of truncation, valuation of data points corresponds to the effectiveness of those data points in model's performance. The second plot shows the rank correlation between the valuation of that level of truncation and the valuation without any truncation. As it is seen, all all cases, values derived by truncation level of $25\%$ have rank correlation around 0.8 with that of having no truncation.

    \paragraph{G-Shapley values and TMC-Shapley values coherency}

        One important question is how much the values returned by G-Shapley are similar. In this sectio we will report the values for some of the experiments presented in Section 4:
        
        \begin{itemize}
        
            \item \textbf{Synthetic datasets} For all data sets in section 4.2 and train set size of 100, using logistic regression model, G-Shapley and TMC-Shapley have correlation coefficient between 0.9 and 0.95. Changing the model to neural network reduces correlation coefficient to the range of 0.7 to 0.8. Increasing the train size to 1000, for the logistic regression model, all correlation coefficients are between 0.94 and 0.97 which for the neural network model reduces to the range of 0.8 to 0.88.
            
            \item \textbf{Disease prediction} For the breast cancer data, G-Shapley and TMC-Shapley have a correlation coefficient of 0.86. For the task of skin cancer prediction, the values of G-Shapley and TMC-Shapley have correlation coefficient of 0.77 and rank correlation of 0.78.  
            
            \item \textbf{Label flip} For the Fashion MNIST data set G-Shapley and TMC-Shapley have correlation coefficient equal 0.62 and for the flower classification data set, the correlation is 0.57.
            
        \end{itemize}

\end{document}